\documentclass[letterpaper, 10 pt, conference]{ieeeconf}  

\IEEEoverridecommandlockouts 
\usepackage{times}
\usepackage{epsfig}
\usepackage{graphicx}
\usepackage{amsmath}
\usepackage{amssymb}
\usepackage{balance}
\usepackage[percent]{overpic}
\usepackage{pifont}
\usepackage{wasysym}
\usepackage{booktabs}

\title{\LARGE {\bf Dilated Point Convolutions:} On the Receptive Field Size \\of Point Convolutions on 3D Point Clouds
\author{Francis Engelmann$^1$, Theodora Kontogianni$^1$, Bastian Leibe$^1$}
\thanks{\vspace{-15pt} \hrule \vspace{3pt}1: All authors are with the Computer Vision Group, Visual Computing Institute at RWTH Aachen University in Aachen, Germany.
}
}

\def\eg{\emph{e.g.}\,}
\def\ie{\emph{i.e.}\,}

\def\etal{\emph{et~al.}\,}
\usepackage{graphicx}
\usepackage[export]{adjustbox}
\usepackage{xcolor}
\usepackage{overpic}
\usepackage{multirow}

\newcommand{\refsec}[1]{Section~\ref{sec:#1}}

\newcommand{\reffig}[1]{Figure~\ref{fig:#1}}
\newcommand{\reftab}[1]{Table~\ref{tab:#1}}
\newcommand{\parag}[1]{\vspace{0ex} \textit{#1}}

\definecolor{M_SemSeg}{RGB}{255, 225, 200}
\definecolor{M_Class}{RGB}{192, 241, 244}
\definecolor{M_PointConv}{RGB}{202, 240, 208}
\newcommand{\colorsquare}[1]{{\color{#1}$\blacksquare$}\hspace{-2.2mm}$\square$}

\newcommand{\legend}[2]{{\color{#1_#2}$\CIRCLE$}\,#2\,\,}
\definecolor{SN_Wall}{RGB}{174, 199, 232}
\definecolor{SN_Floor}{RGB}{152, 223, 138}
\definecolor{SN_Cabinet}{RGB}{31, 119, 180}
\definecolor{SN_Bed}{RGB}{255, 187, 120}
\definecolor{SN_Chair}{RGB}{188, 189, 34}
\definecolor{SN_Sofa}{RGB}{140, 86, 75}
\definecolor{SN_Table}{RGB}{255, 152, 150}
\definecolor{SN_Door}{RGB}{214, 39, 40}
\definecolor{SN_Window}{RGB}{197, 176, 213}
\definecolor{SN_Bookshelf}{RGB}{148, 103, 189}
\definecolor{SN_Picture}{RGB}{196, 156, 148}
\definecolor{SN_Counter}{RGB}{23, 190, 207}
\definecolor{SN_Desk}{RGB}{247, 182, 210}
\definecolor{SN_Curtain}{RGB}{219, 219, 141}
\definecolor{SN_Refrigerator}{RGB}{255, 127, 14}
\definecolor{SN_Showercurtain}{RGB}{158, 218, 229}
\definecolor{SN_Toilet}{RGB}{44, 160, 44}
\definecolor{SN_Sink}{RGB}{112, 128, 144}
\definecolor{SN_Bathtub}{RGB}{227, 119, 194}
\definecolor{SN_Otherfurn.}{RGB}{82, 84, 163}

\begin{document}

\maketitle
\thispagestyle{empty}
\pagestyle{empty}

\begin{abstract}
In this work, we propose \emph{Dilated Point Convolutions} (DPC).
In a thorough ablation study, we show that the receptive field size is directly related to the performance of 3D point cloud processing tasks,
including semantic segmentation and object classification. 
Point convolutions are widely used to efficiently process 3D data representations such as point clouds or graphs.
However, we observe that the receptive field size of recent point convolutional networks is inherently limited.
Our dilated point convolutions alleviate this issue, they significantly increase the receptive field size of point convolutions.
Importantly, our dilation mechanism can easily be integrated into most existing point convolutional networks.
To evaluate the resulting network architectures, we visualize the receptive field and report competitive scores on popular point cloud benchmarks.
\end{abstract}


\section{Introduction}
\label{sec:intro}
The past years have witnessed a tremendous development of 3D scene understanding methods on several tasks
including semantic segmentation\,\cite{Qi17CVPR}, object detection\,\cite{Zhou18CVPR}, and instance segmentation\,\cite{Elich19GCPR}.
Recent advancements such as point convolutional layers \cite{Wang18CVPRa, Wang18CoRR, Wu18CVPR} which can directly operate on 3D point clouds further boosted the field.

In the 2D image domain, analyzing the receptive field is an important tool for diagnosing and comprehending convolutional neural networks (CNN).
The receptive field of a neural unit describes the region of the input data that influences its output value.
All input data outside of the receptive field does not contribute to the output.
Hence, large receptive fields are important since they enable reasoning on a larger input context.

Current successful architectures operating on grid-like data (\eg images \cite{Simonyan15ICLR, Szegedy15CVPR, He16CVPR}), increase the receptive field implicitly by using \emph{deeper} network architectures.
However, only few works explicitly study the influence of receptive fields in the domain of 2D image CNNs \cite{Luo16NIPS, Mishkin17CVIU}.
So far, there is no work analyzing the receptive fields of deep networks operating directly on 3D point clouds.
Such a study is particularly challenging, since the theoretical size of receptive fields is difficult to compute due to the non-uniform structure of 3D point clouds.
Nevertheless, we argue that the concept of receptive fields is equally important in the 3D domain.

Point convolutional layers \cite{Wang18CVPRa, Wang18CoRR, Wu18CVPR} are a major driving force behind the success of networks that can directly operate on unstructured data such as 3D point clouds.
Furthermore, they can be seen as a generalization of \emph{discrete} convolutions.
While continuous point convolutions operate on data sampled at continuous positions in space,
discrete convolutions operate on grid-structured data such as images or voxel-grids, \ie the data is sampled at discrete positions.

As such, we propose to visualize the receptive fields to analyze different network architectures and we present a thorough ablation study comparing several strategies which increase the receptive field of point convolutions.
Specifically, we look at common strategies to increase the receptive field by 1) stacking convolutional layers and 2) using larger kernel sizes.
By visually analyzing the extent of the resulting receptive fields, we notice that their influence still remains rather limited.
Motivated by these observations, we propose \emph{Dilated Point Convolutions} as a means to significantly increase the receptive field size of point convolutions.

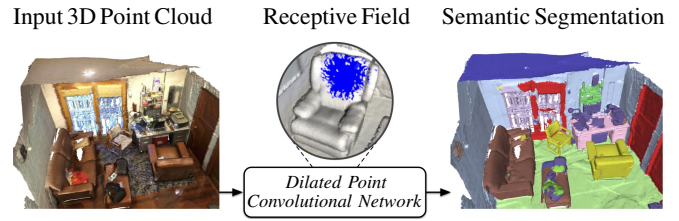
\begin{figure}[t]
\vspace{20px}
\begin{overpic}[unit=1mm,width=\columnwidth]{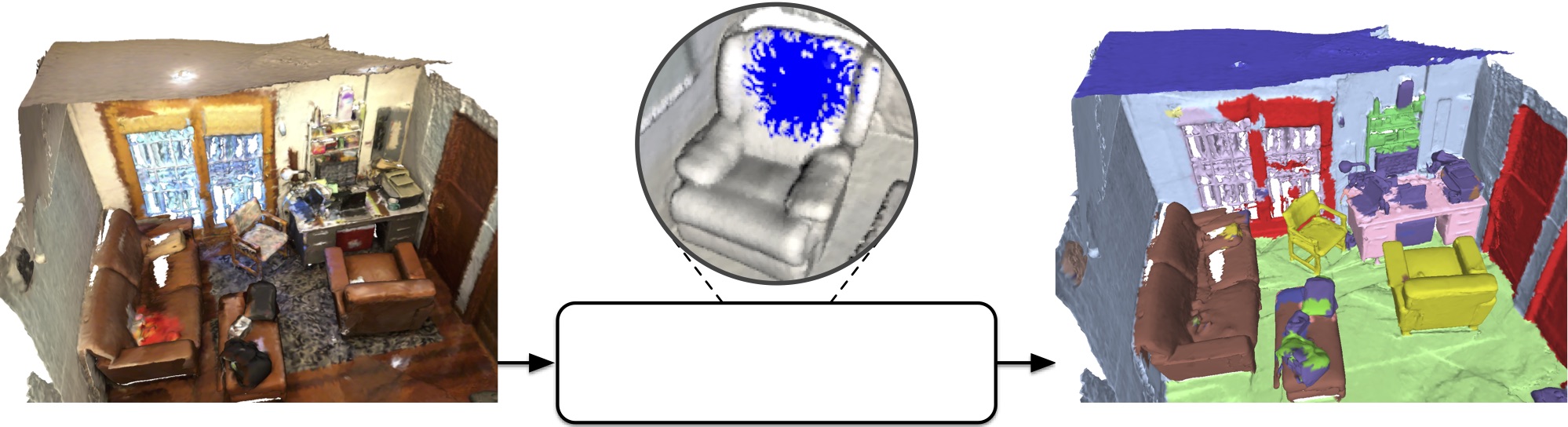}
{\small
\put(0,30){Input\,3D\,Point\,Cloud}
\put(38.5,30){Receptive\,Field}
\put(66,30){Semantic\,Segmentation}
}{\scriptsize
\put(42,4.7){\textit{Dilated Point}}
\put(36,1.7){\textit{Convolutional Network}}
}
\end{overpic}
\vspace{-15px}
\caption{This work presents \emph{Dilated Point Convolutions} (DPC).
We observe that existing point convolutional networks have inherently small receptive field sizes.
Assisted by this observation, we compare different network architectures and propose our dilation mechanism as a simple yet elegant solution to significantly increase the receptive field size of point convolutions and improve their performance on multiple point cloud processing tasks.
}
\label{fig:teaser}
\end{figure}

The paper is structured as follows: We start by discussing current methods for 3D point cloud processing and existing works analyzing receptive fields on discrete convolutions. Then, we review \emph{Point Convolutions} as an instance of continuous convolutions on 3D point clouds. Next, we describe and visualize well established methods for increasing receptive fields, which leads us to the derivation of \emph{Dilated Point Convolutions}. Finally, in the experimental section, we compare the aforementioned strategies.

Our contributions are as follows:
(1) We evaluate most commonly used strategies to increase the receptive fields in current methods using point convolutions.
(2) We propose to visualize the receptive field of point convolutions to make educated network design choices.
(3) From these observations, we derive \emph{Dilated Point Convolutions} (DPC) as an elegant mechanism to significantly increase the receptive field size.
(4) Using DPCs we are able to report competitive scores on the task of 3D semantic segmentation on S3DIS\,\cite{Armeni16CVPR} and ScanNet\,\cite{Dai17CVPR} as well as shape classification on ModelNet40\,\cite{Song15CVPR}.

\begin{figure*}[ht]
\center
\vspace{3px}
\rule{0pt}{1ex}
\hspace{2.24mm}
\includegraphics[width=1\textwidth]{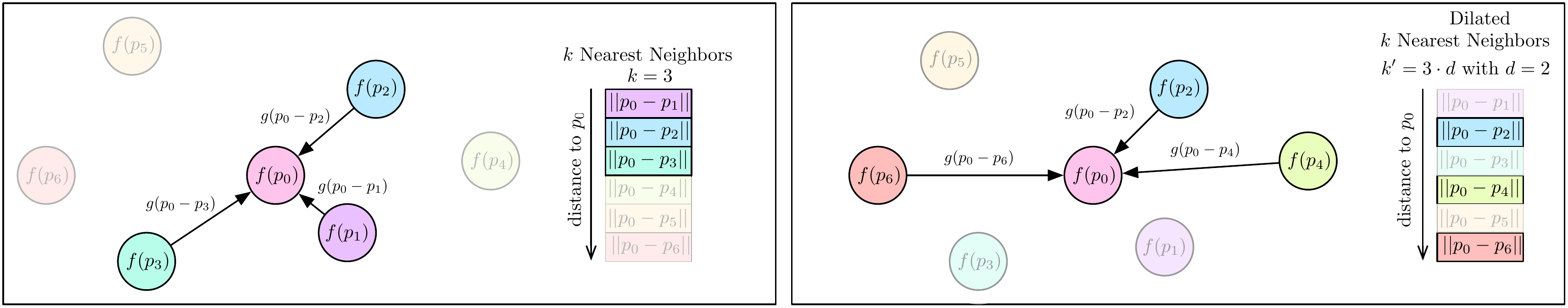}
\vspace{-114px}

\begin{tabular}{cc}
\fcolorbox{black}{gray!10}{\footnotesize\textit{Point Convolutions}}\hspace{20px} &
\fcolorbox{black}{gray!10}{\footnotesize\textit{Dilated Point Convolutions}}\hspace{40px} \\
\hspace{0.5\textwidth} & \hspace{0.5\textwidth} \\
\end{tabular}
\vspace{70px}

   \caption{
   \emph{(Left)} \textbf{Point Convolutions.}
    Schematic illustration of point convolutions.
    The continuous feature function $f(\cdot)$ assigns a feature value to continuous point positions $p$.
    \emph{(Right)}
    \textbf{Dilated Point Convolutions.}
    We propose \emph{dilated} point convolutions as an elegant mechanism to significantly increase the receptive field of point convolutions
    resulting in a notable boost in performance at almost no additional computational cost (see \reftab{segmentation_results_stanford_dilation}).
    Instead of computing the kernel weights $g(\cdot)$ over the $k$ nearest neighbors,
    we propose to compute the kernel weights over a \emph{dilated} neighborhood obtained by computing the sorted $k \cdot d$ nearest neighbors and preserving only every $d$-th point.}
\label{fig:point_convolutions}
\end{figure*}

\section{Related Work}
\label{sec:related}

\parag{2D Projection Representation.}
Qi~\etal \,\cite{Qi16CVPR} and Boulch \etal\,\cite{Boulch17CG} project 3D point clouds into 2D representations, then apply 2D convolutional networks and finally fuse the results back into 3D space.
These type of projections do not make use of the underlying geometric structure as they only operate on the projected appearance of the point clouds.

\parag{3D Volumetric Grid Representation.}
Maturana and Scherer\,\cite{Maturana15IROS} and Song~\etal\cite{Song15CVPR} voxelize point clouds into regular volumetric grids and apply 3D convolutions.
These approaches are constrained by the fixed resolution of the 3D grid.
Coarse grids lead to loss of detail and fine ones suffer from high memory and computational costs.
The use of octrees\,\cite{Riegler18CVPR} and kd-trees\,\cite{Klokov17ICCV} offer improved grid resolutions.
Recently, Graham \etal\,\cite{Graham18CVPR} offered a speed- and memory-efficient approach for sparse 3D convolutions which are applied only on occupied voxels.
However, voxelized point clouds can still be problematic if adjacent points are far apart, which can hinder information flow. 

\parag{3D Feature Learning on Point Sets.}
Numerous methods operate directly on 3D point clouds \cite{Qi17CVPR, Wang18CVPRa, Wang18CoRR, Wu18CVPR}.
They follow-up on the seminal work of \emph{PointNet}\,\cite{Qi17CVPR} which applies point-wise multi-layer-perceptrons (MLP) followed by max-pooling over all points to extract a global point cloud descriptor but fails to capture local structure.
Local structure is implicitly considered in 2D images and 3D voxels by using spatial grids.
Filters that incorporate the information of the neighboring points in the grid are then learned.
Numerous methods rely on similar types of spatial neighborhoods in an unstructured point cloud:
Hua~\etal\,\cite{Hua18CVPR} compute nearest neighbors on the fly and bin them into spatial cells before using fully convolutional networks.
Landrieu and Boussaha\,\cite{Landrieu19CVPR} compute neighborhoods by over-segmenting 3D point clouds into superpoints.
However, the most popular method used by \cite{Engelmann18ECCVW, Li18NIPS, Qi17NIPS, Wang18CoRR, Wu18CVPR} consists in computing the $k$ nearest neighbors (KNN) of every point to represent its neighborhood.
\emph{EdgeConvs}\,\cite{Wang18CoRR} establish this neighborhood on the feature space while \emph{PointConv}\,\cite{Wu18CVPR}, \emph{PointNet++}\,\cite{Qi17NIPS} and  \emph{PointCNN}\,\cite{Li18NIPS} use the spatial coordinates.
Engelmann \etal\,\cite{Engelmann18ECCVW} use KNN in the feature space and k-means in the world coordinate system to create neighborhoods.

\parag{Receptive Field Analysis.} Few works systematically study the influence of receptive fields on 2D image CNNs\,\cite{Luo16NIPS, Mishkin17CVIU}.
In general, deeper networks which stack multiple layers of 2D convolutions have proven to work better \cite{Simonyan15ICLR, Szegedy15CVPR}.
Dilated convolutions~\cite{Yu16ICLR}, previously introduced as \emph{atrous} convolutions~\cite{Chen15ICLR}, used in 2D image semantic segmentation, allow to efficiently enlarge the receptive field of filters to incorporate larger context without increasing the number of model parameters.
In this work, we propose a simple yet effective dilation mechanism for 3D point convolutions.

\begin{figure*}
\begin{center}
\rule{0pt}{1ex}
\hspace{2.24mm}
\includegraphics[width=\textwidth]{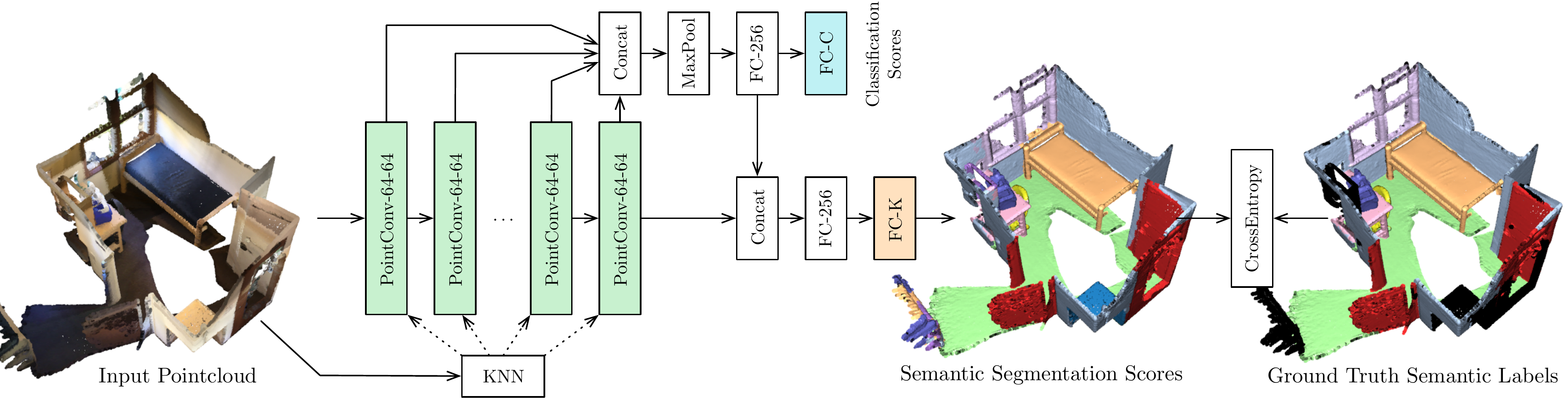}
\vspace{-20px}
\end{center}
   \caption{Our model is built from a sequence of \emph{point convolutional layers} (PointConv\,\colorsquare{M_PointConv}, \refsec{dilated_convolutions}).
   Fully connected layers are denoted by FC.
   The bottom output branch \colorsquare{M_SemSeg} is used for the experiments on semantic segmentation.
   The top output branch \colorsquare{M_Class} is used for object classification.
   Each task is supervised using a cross-entropy loss, with either $K$ semantic classes for semantic segmentation or $C$ object classes for the object classification task.}
\label{fig:model}
\end{figure*}
\section{Approach}
\label{sec:method}
In this section, we formally define \emph{point convolutions} and examine the importance of a large \emph{receptive field size} in the context of 3D point cloud processing.
We revisit existing strategies to increase the receptive field.
Then, we propose our main contribution \emph{dilated point convolutions}, an elegant yet easy technique to significantly increase the receptive field size of point convolutional networks.

\subsection{Point Convolutions}
Point convolutions can be formulated using the general definition of continuous convolutions in a $D$-dimensional space.
Continuous convolutions are defined as
\begin{equation}
	\big(f * g \big) (p_i) = \int_{-\infty}^{+\infty} f(p_j) \odot g(p_i - p_j)\, \mathop{dp_j},
\end{equation}
where $\odot$ is the Hadamard-product of the continuous feature function $f : \mathbb{R} ^D \rightarrow \mathbb{R} ^F$ assigning a feature-vector $f(p_j) \in \mathbb{R}^F$ to each position $p_j \in \mathbb{R}^D$,
and the continuous kernel function $g : \mathbb{R} ^D \rightarrow \mathbb{R}^F$ mapping a relative position to a kernel weight.
In the case of 3D point clouds, we have $D=3$ and the feature vector could for example contain the point position, color, and normal such that $f(p) \in \mathbb{R}^9$ , see \reffig{point_convolutions}.
In most practical applications, \eg~reconstructed 3D point clouds,  the feature function $f$ is not fully known since only a limited number $N$ of point positions $p_n$ are observed or even occupied.
Using Monte-Carlo integration, the continuous convolution can then be approximated as
\begin{equation}
\big(f * g \big) (p_i) \approx \frac{1}{N} \sum_{n=1}^N f(p_n) \odot g(p_i - p_n),
\end{equation}
where recent methods implement the kernel function $g(\cdot)$ as a learned parametric function based on a multi-layer perceptron (MLP)
\begin{equation}
g(p; \theta) = \text{MLP}(p\,; \theta),
\end{equation}
where $p$ is the relative position between two points and $\theta$ is a set of learned parameters.
In order to extract high-frequency signals it is important to define localized kernels \cite{Zeiler18ECCV}.
In 2D image CNNs, this is implemented by \eg $3 \times 3$ or $5 \times 5$ pixel kernels.
For point convolutions, this effect is achieved by limiting the cardinality of the local kernel support, \ie by defining a local neighborhood $\mathcal{N}_i$ around each point $p_i$
\begin{equation}
\big(f * g \big) (p_i) \approx \frac{1}{|\mathcal{N}_i|} \sum_{p_k \in \mathcal{N}_i} f(p_k) \odot g(p_i - p_k).
\end{equation}

The above definition of continuous convolutions is used in Wang \etal\,\cite{Wang18CVPRa} and 
\emph{PointConv}\,\cite{Wu18CVPR} which additionally proposes to weight the kernel function using the inverse local density to compensate for the non-uniform distribution of point samples.
In \emph{SpiderCNN}, Xu \etal \,\cite{Xu18ECCV} propose to replace the MLP by a combination of step functions and Taylor expansions to capture rich spatial information.
A broader interpretation of continuous convolutions is used in \emph{EdgeConv}\,\cite{Wang18CoRR}, where the kernel function $g(\cdot)$ is not only defined over relative positions but also over the difference of learned point features.
Independent of the concrete implementation, all previously mentioned methods, including \emph{PointCNN}\,\cite{Li18NIPS}, rely on $k$ nearest neighbors (KNN) to define a local neighborhood $\mathcal{N}$ resulting in local kernels.
Next, after looking at the receptive field size, we use KNN neighborhoods to define \emph{dilated} point convolutions.

\subsection{Receptive Field Size.}
\begin{figure*}
\center
\small
\begin{tabular}{ccccccc}
	 &\textit{1 Point Conv} & \textit{2 Point Convs} & \textit{3 Point Convs} & \textit{5 Point Convs} & \textit{7 Point Convs}\\
	\vspace{-0.25cm}
	&\hspace{0.17\textwidth} & \hspace{0.17\textwidth} & \hspace{0.17\textwidth} & \hspace{0.17\textwidth} & \hspace{0.17\textwidth} 
\end{tabular}
	
\rotatebox[origin=l]{90}{\hspace{20px}$k=5$, $d=1$}
\includegraphics[width=0.19\linewidth, trim={80 60 40 60}, clip, frame]{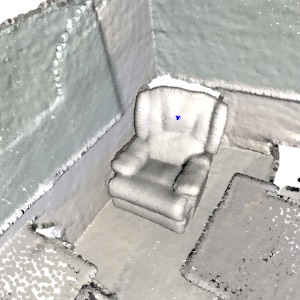}\hspace{2px}%
\includegraphics[width=0.19\linewidth, trim={80 60 40 60}, clip, frame]{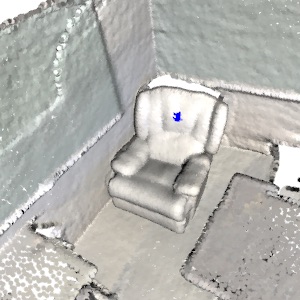}\hspace{2px}%
\includegraphics[width=0.19\linewidth, trim={80 60 40 60}, clip, frame]{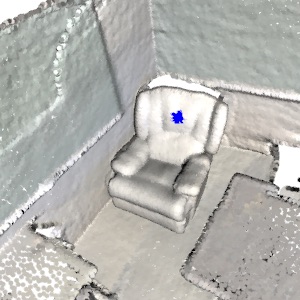}\hspace{2px}%
\includegraphics[width=0.19\linewidth, trim={80 60 40 60}, clip, frame]{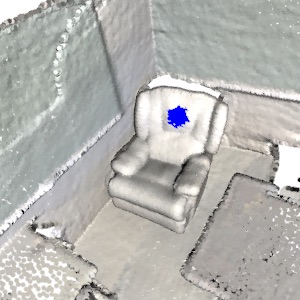}\hspace{2px}%
\includegraphics[width=0.19\linewidth, trim={80 60 40 60}, clip, frame]{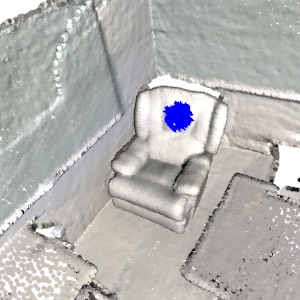}\\
\vspace{2px}
\rotatebox[origin=l]{90}{\hspace{16px}$k=10$, $d=1$}
\includegraphics[width=0.19\linewidth, trim={80 60 40 60}, clip, frame]{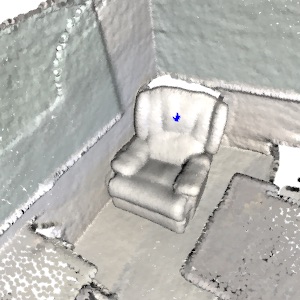}\hspace{2px}%
\includegraphics[width=0.19\linewidth, trim={80 60 40 60}, clip, frame]{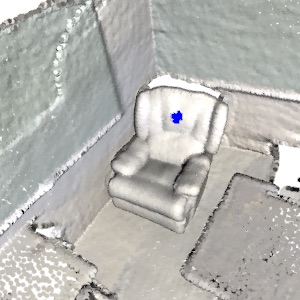}\hspace{2px}%
\includegraphics[width=0.19\linewidth, trim={80 60 40 60}, clip, frame]{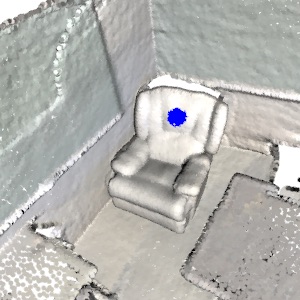}\hspace{2px}%
\includegraphics[width=0.19\linewidth, trim={80 60 40 60}, clip, frame]{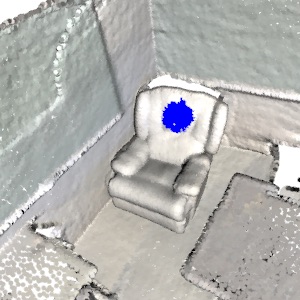}\hspace{2px}%
\includegraphics[width=0.19\linewidth, trim={80 60 40 60}, clip, frame]{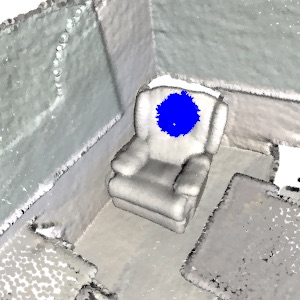}\\
\vspace{2px}
\rotatebox[origin=l]{90}{\hspace{18px}$k=20$, $d=1$}
\includegraphics[width=0.19\linewidth, trim={80 60 40 60}, clip, frame]{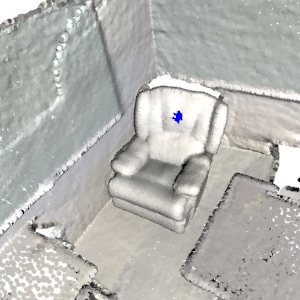}\hspace{2px}%
\includegraphics[width=0.19\linewidth, trim={80 60 40 60}, clip, frame]{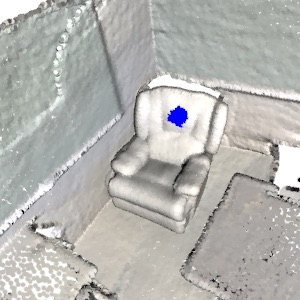}\hspace{2px}%
\includegraphics[width=0.19\linewidth, trim={80 60 40 60}, clip, frame]{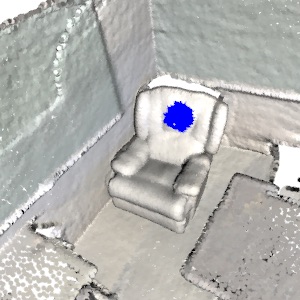}\hspace{2px}%
\includegraphics[width=0.19\linewidth, trim={80 60 40 60}, clip, frame]{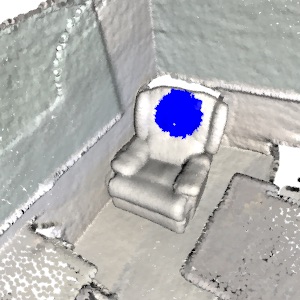}\hspace{2px}%
\includegraphics[width=0.19\linewidth, trim={80 60 40 60}, clip, frame]{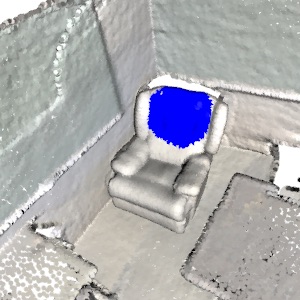}\\
\vspace{2px}
\rotatebox[origin=l]{90}{\hspace{18px}$k=20$, $d=2$}
\includegraphics[width=0.19\linewidth, trim={80 60 40 60}, clip, frame]{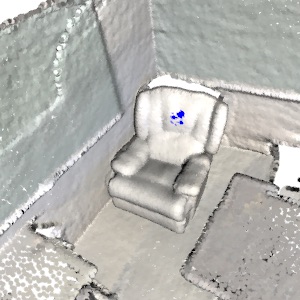}\hspace{2px}%
\includegraphics[width=0.19\linewidth, trim={80 60 40 60}, clip, frame]{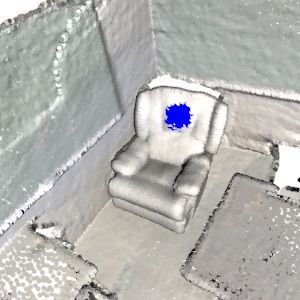}\hspace{2px}%
\includegraphics[width=0.19\linewidth, trim={80 60 40 60}, clip, frame]{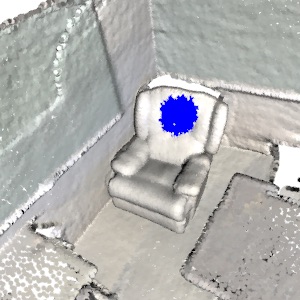}\hspace{2px}%
\includegraphics[width=0.19\linewidth, trim={80 60 40 60}, clip, frame]{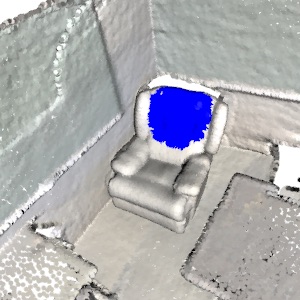}\hspace{2px}%
\includegraphics[width=0.19\linewidth, trim={80 60 40 60}, clip, frame]{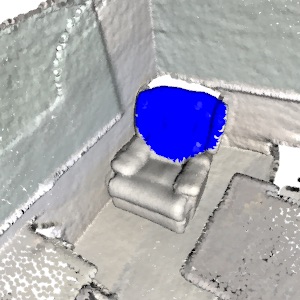}\\
\vspace{2px}
\rotatebox[origin=l]{90}{\hspace{18px}$k=20$, $d=8$}
\includegraphics[width=0.19\linewidth, trim={80 60 40 60}, clip, frame]{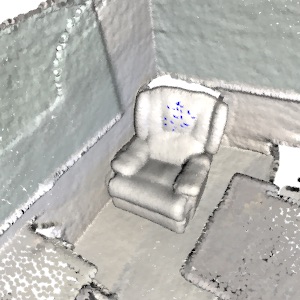}\hspace{2px}%
\includegraphics[width=0.19\linewidth, trim={80 60 40 60}, clip, frame]{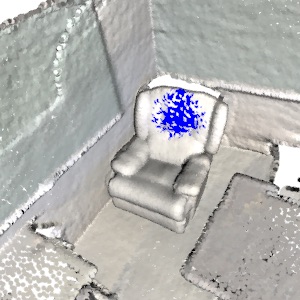}\hspace{2px}%
\includegraphics[width=0.19\linewidth, trim={80 60 40 60}, clip, frame]{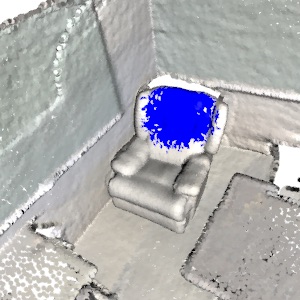}\hspace{2px}%
\includegraphics[width=0.19\linewidth, trim={80 60 40 60}, clip, frame]{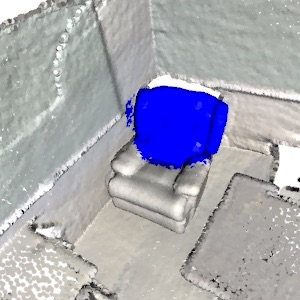}\hspace{2px}%
\includegraphics[width=0.19\linewidth, trim={80 60 40 60}, clip, frame]{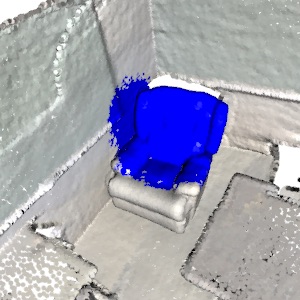}\\
\vspace{2px}
\rotatebox[origin=l]{90}{\hspace{15px}$k=20$, $d=16$}
\includegraphics[width=0.19\linewidth, trim={80 60 40 60}, clip, frame]{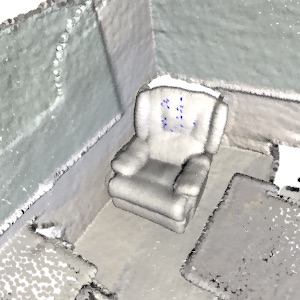}\hspace{2px}%
\includegraphics[width=0.19\linewidth, trim={80 60 40 60}, clip, frame]{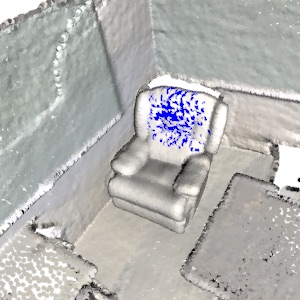}\hspace{2px}%
\includegraphics[width=0.19\linewidth, trim={80 60 40 60}, clip, frame]{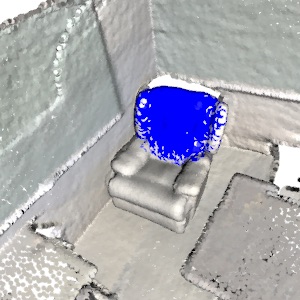}\hspace{2px}%
\includegraphics[width=0.19\linewidth, trim={80 60 40 60}, clip, frame]{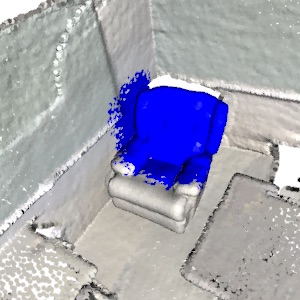}\hspace{2px}%
\includegraphics[width=0.19\linewidth, trim={80 60 40 60}, clip, frame]{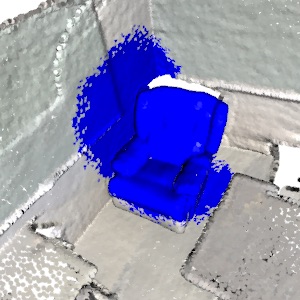}
\caption{\textbf{Receptive field} visualized in blue for different network architectures using an increasing number of \emph{Point Convolutions} (columns) and increasing kernel sizes (rows) based on the number of nearest neighbors $k$ and dilation factor $d$.
The receptive field sizes of point convolutions without dilation ($d=1$) are substantially smaller.
However, for large dilations, \eg $d=16$ the receptive field is sparse in early stages of the deep network (bottom left).}
\label{fig:receptive_field}
\end{figure*}

A large receptive field is directly related to the performance of point convolutional networks (\refsec{experiments}).
Thus our goal is to increase the size of the receptive field.
The receptive field (or \emph{field of view}) of a neural unit within a deep network describes the region of the input point cloud that influences the output of that particular unit.
In the context of 3D semantic segmentation, where the task is to assign a semantic label to each point in a given point cloud, the final decision on the label for a particular point is influenced only by those points which lie inside the receptive field of that particular point.
All other points outside the receptive field do not contribute to the decision, see \reffig{receptive_field}.
It is thus essential to design architectures with receptive fields large enough to cover the necessary context for each point.

A common approach to increase the receptive field size, similar to 2D architectures, consists in \emph{stacking} multiple (point) convolutional layers.
\emph{EdgeConvs}\,\cite{Wang18CoRR} stack 3 convolutional layers, \emph{SpiderCNN}\,\cite{Xu18ECCV} use 4 layers and \emph{PCCN}\,\cite{Wang18CVPRa} use 8.
Here, we compare 3, 5 and 7 layers, see \reftab{segmentation_results_stanford_depth}.

Increasing the \emph{kernel size} of the convolution is another popular technique.
In the setup of point convolutions this effect is achieved by selecting a larger number $k$ of nearest neighbors.
Note, however, that this does not increase the number of model parameters since the kernel weights are computed over relative point positions using the parametric kernel function $g(\cdot)$, see \reftab{segmentation_results_stanford_depth}.
This is in stark contrast to convolutions defined over discrete grid positions (\eg 2D image CNN) where a larger kernel increases the number of model parameters.

\subsection{Dilated Point Convolutions.}
\label{sec:dilated_convolutions}
Using the previously mentioned approaches, the receptive field size still remains limited, see top 3 rows in \reffig{receptive_field}.
Therefore, we propose \emph{dilated point convolutions} (DPC) as an elegant yet efficient mechanism to increase the receptive field size.
DPCs are equal to point convolutions (PC), however, they differ in the way they select neighboring points:
While PCs directly use the $k$ nearest neighbors, DPCs first compute the $k \cdot d$ nearest neighbors and then select every $d$-th neighbor, see \reffig{point_convolutions} (right).
Note that for $d\,$=$\,1$, DPCs are identical to PCs.
The dilation causes a significantly increased receptive field size (see \reffig{receptive_field}).
However, the number of parameters remains unchanged.
The larger number $k \cdot d$ of neighbors that needs to be computed adds a sublinear computational overhead. See \reftab{segmentation_results_stanford_dilation}.
Another positive aspect about DPC is that they can directly be added -- with minimal modifications -- to most existing point convolutional networks, if the local kernel neighborhood $\mathcal{N}$ originates from a nearest neighbor search.

\begin{figure*}[htp]
\center
\vspace{5pt}
\includegraphics[width=0.245\linewidth, trim={0 20  0 0}, clip, frame]{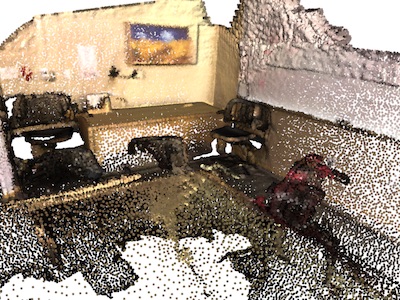}\hspace{2px}%
\includegraphics[width=0.245\linewidth, trim={0 20  0 0}, clip, frame]{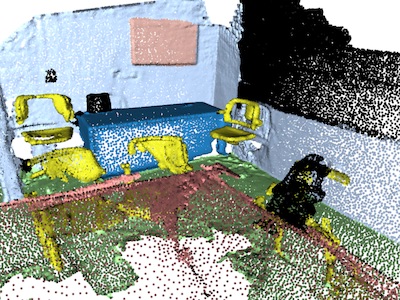}\hspace{2px}%
\includegraphics[width=0.245\linewidth, trim={0 20  0 0}, clip, frame]{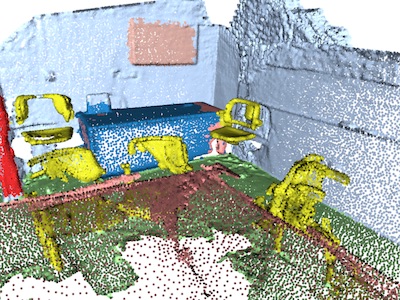}\hspace{2px}%
\includegraphics[width=0.245\linewidth, trim={0 20  0 0}, clip, frame]{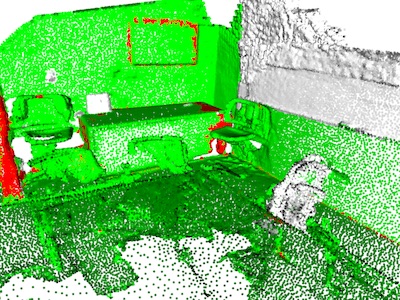}\\
\vspace{2pt}
\includegraphics[width=0.245\linewidth, trim={0 20  0 0}, clip, frame]{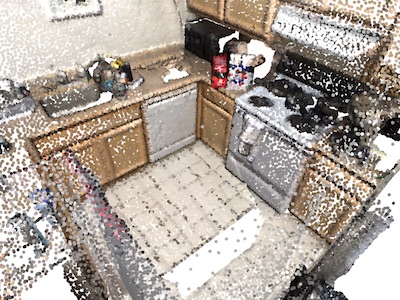}\hspace{2px}%
\includegraphics[width=0.245\linewidth, trim={0 20  0 0}, clip, frame]{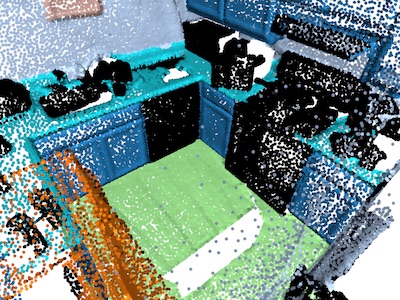}\hspace{2px}%
\includegraphics[width=0.245\linewidth, trim={0 20  0 0}, clip, frame]{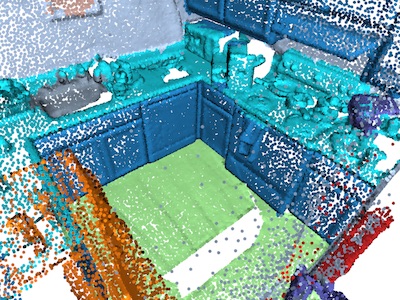}\hspace{2px}%
\includegraphics[width=0.245\linewidth, trim={0 20  0 0}, clip, frame]{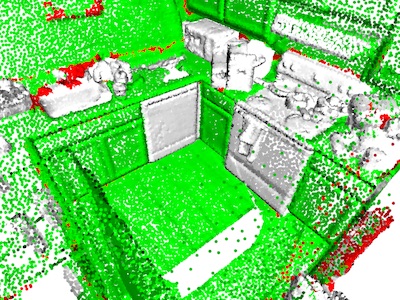}\\
\vspace{2pt}
\includegraphics[width=0.245\linewidth, trim={0 0  0 20}, clip, frame]{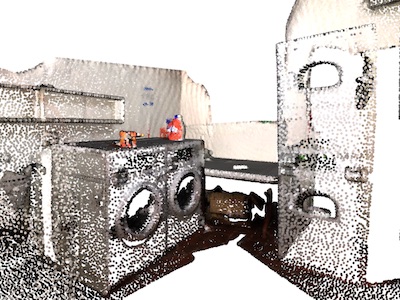}\hspace{2px}%
\includegraphics[width=0.245\linewidth, trim={0 0  0 20}, clip, frame]{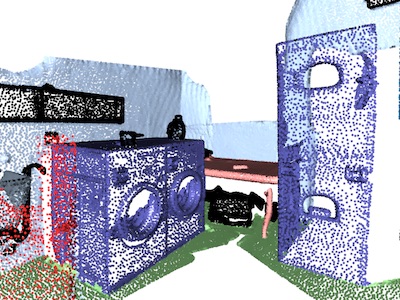}\hspace{2px}%
\includegraphics[width=0.245\linewidth, trim={0 0  0 20}, clip, frame]{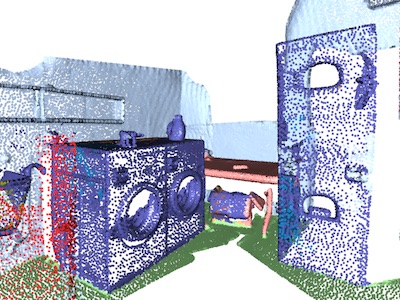}\hspace{2px}%
\includegraphics[width=0.245\linewidth, trim={0 0  0 20}, clip, frame]{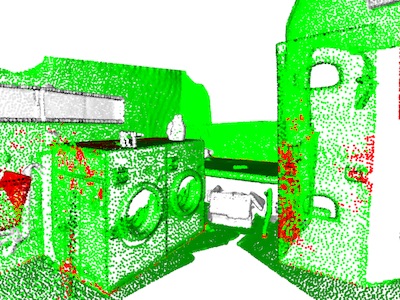}

\begin{small}
\vspace{-8px}
\begin{tabular}{cccc}
\fcolorbox{black}{gray!10}{\textit{Point Cloud}} &
\fcolorbox{black}{gray!10}{\textit{Ground Truth}} &
\fcolorbox{black}{gray!10}{\textit{Our Predictions}} &
\fcolorbox{black}{gray!10}{\textit{Error}}\\
\hspace{0.22\textwidth} & \hspace{0.22\textwidth} & \hspace{0.22\textwidth} & \hspace{0.22\textwidth} \\
\end{tabular}
\\
\legend{SN}{Wall}
\legend{SN}{Floor}
\legend{SN}{Cabinet}
\legend{SN}{Chair}
\legend{SN}{Table}
\legend{SN}{Door}
\legend{SN}{Picture}
\legend{SN}{Counter}
\legend{SN}{Desk}
\legend{SN}{Refrigerator}
\legend{SN}{Sink}
\legend{SN}{Otherfurn.}
\end{small}

\caption{\textbf{Results of our method on ScanNet\,v2 dataset \cite{Dai17CVPR} validation.} Left to right: Input RGB point cloud, semantic segmentation ground truth, semantic segmentation prediction and the error where green shows correct predictions, red shows wrong predictions and white indicates unlabeled ground truth.}
\label{fig:scannet_quali}
\end{figure*}


\section{Experiments}
\label{sec:experiments}

\parag{Model Architecture.}
In all our experiments, we use a deep convolutional model as depicted in \reffig{model}.
The main branch (shown in green) consists of stacked (dilated) point convolutions.
The $k$ nearest neighbors (KNN) for each point are computed on-the-fly.
The final point features are concatenated with global features obtained by max-pooling over the concatenated point features at different depth-levels.

\subsection{3D Semantic Segmentation}
\parag{Task and Metrics.}
The goal is to predict a semantic label for each point in a given point cloud.
This task is especially well-suited to analyze the effectiveness of larger receptive fields, since the label of each point is only influenced by points in its receptive field.
We adopt the commonly used metrics: mean intersection over union (mIoU), mean class accuracy (mAcc), and overall accuracy (oAcc).

\parag{Datasets.}
We evaluate on two datasets:
(1) Stanford Large-Scale 3D Indoor Spaces (S3DIS)\,\cite{Armeni16CVPR} contains dense 3D point clouds from 6 large-scale indoor areas, consisting of 271 rooms from 3 different buildings.
The points are annotated with 13 semantic classes.
We use the common train/test split, which trains on all areas except \emph{Area 5} which we keep for testing\,\cite{Armeni16CVPR, Tchapmi173DV, Wang18CVPRa}.
(2) {ScanNet\,v2\,\cite{Dai17CVPR}}
contains 3D scans of a wide variety of indoor scenes, including apartments, hotels, conference rooms and offices.
The dataset contains 20 valid semantic classes.
We use the public training, validation and test split of 1201, 312 and 100 scans, respectively.

\parag{Training Details.}
We train our networks using the Adam optimizer and exponential-decay learning-rate scheduling.
During training we randomly sample 4092 points from crops of 3\,m side length.
This differs from most concurrent methods which train on 1\,m or 1.5\,m crops.
Since our model has a much larger receptive field it can learn to make use of this additional context.  
In general, small training crops could hinder the network to learn from larger context as soon as the size of the receptive field exceeds the size of the training crops.
Points are sampled without replacement and we use zero-padding if there are less than 4092 points.

\parag{Results and Discussion}
\label{sec:results}
We report scores of our best performing models on the  ScanNet\,v2 dataset\,\cite{Dai17CVPR} and the S3DIS dataset\,\cite{Armeni16CVPR} in \reftab{segmentation_results_stanford}.
\begin{table}[b]
\center
\caption{3D Semantic segmentation on S3DIS (A5) and ScanNet\,V2.}
\setlength{\tabcolsep}{10pt}
\begin{tabular}{crcccc}
\toprule 
& \textbf{Method} & & \textbf{mIoU} & \textbf{mAcc} & \textbf{oAcc} \\
\midrule
\multirow{6}{*}{\rotatebox{90}{\tiny S3DIS Area\,5}}
& PointNet\,\cite{Qi17CVPR} 								& & 41.1 & 49.0 & -\\
& KWYND\,\cite{Engelmann18ECCVW} 				& & 52.2&  59.1 & 84.2\\
& PointCNN\,\cite{Li18NIPS}								& & 57.3 & 63.9 & 85.9 \\
& SPG\,\cite{Landrieu18CVPR}							& & 58.0 & 66.5 & 86.4 \\
& PCNN\,\cite{Wang18CVPRa}							& & 58.3 & 67.0 & -\\
& \textbf{DPC (Ours)}			 & &  \textbf{61.28} & \textbf{68.38} & \textbf{86.78} \\
\midrule 
\multirow{2}{*}{\rotatebox{90}{\tiny ScanNet}}& DPC (Val-set)		& 								& 59.52 & 67.21 & 85.95 \\
& DPC (Test-set)									& & 59.2 & - & - \\
\bottomrule
\end{tabular} 
\label{tab:segmentation_results_stanford}    
\end{table}
Our dilated point convolutional model is able to outperform other recent KNN-based point convolutional networks by a significant margin on S3DIS, and provides competitive scores on ScanNet, specifically among point convolutional approaches.
In \reffig{scannet_quali}, we show qualitative results on the ScanNet validation dataset.
We highlight wrong predictions in red (see right-most column).

\subsection{3D Object Classification}
\parag{Dataset.} ModelNet40 consists of CAD models that belong to one of 40
different categories. We use the official split of 9843 shapes
for training and 2468 for testing, as in \cite{Qi17NIPS}.
We randomly sample 4,000 points from the 3D model of an object.
The input features are the 3D coordinates and the surface normals (6 input channels in total).
\parag{Comparison.}
Table \ref{tab:scores_modelnet40} shows the comparison between our method and prior methods.
We report overall classification accuracy (oAcc) and mean classification accuracy (mAcc).
Next, we present an ablation study of all model hyper-parameters.
\begin{table}[t]
\center
\vspace{5px}
\caption{Object classification scores on ModelNet40}
\setlength{\tabcolsep}{9pt}
\begin{tabular}{rccc}
\toprule 
\textbf{Method} & \textbf{\#\,Points} &\textbf{oAcc} & \textbf{mAcc}   \\
\midrule
PointNet\cite{Qi17CVPR}								& 1k 		& 89.2 & 86.2\\
PointNet++(with normals)\cite{Qi17NIPS}		& 5k 		& 91.9 & -\\
Kd-Net\cite{Klokov17ICCV}							& 32k 	& 91.8 &88.5\\
EdgeConv\cite{Wang18CoRR}						& 1k		& 92.2 & 90.2\\
SO-Net(with normals)\cite{Li18CVPR}			& 5k		& 92.4 & 90.8\\
SpiderCNN(with normals)\cite{Xu18ECCV}	& 1k		& 92.4 & -\\
\textbf{DPC (Ours)} with normals 					& 4k		& \textbf{93.1} & \textbf{91.4}   \\
\bottomrule 
\vspace{-17px}
\end{tabular} 
\label{tab:scores_modelnet40}    
\end{table}

\subsection{Ablation Study}
We perform an ablation study on the previously introduced mechanisms for increasing the receptive field size.
The hyper-parameters that we analyze in particular are the number of point convolutional layers, the nearest neighbors $k$ and the dilation factor $d$.
The main results are presented in 
\reftab{segmentation_results_stanford_depth} and \reftab{segmentation_results_stanford_dilation}.
The ablation studies are performed on Area 5 of the S3DIS dataset\,\cite{Armeni16CVPR}.
In the following, we discuss the influence of the individual parameters.

\parag{Depth and Number of Neighbors \textit{k} (\reftab{segmentation_results_stanford_depth}).}
\begin{table}[b]
\center
\caption{Ablation study: stacking point convolutions and varying kernel size $k$. Dataset: S3DIS Area 5.
}
\resizebox{0.99\columnwidth}{!}
{%
\setlength{\tabcolsep}{3pt}
\begin{tabular}{cccccc}
\toprule
{Number of}    & {Number of}     & {Time per    }  & {Number of}  &      &\\
{PointConvs} & {Neighbors $k$} & {Forward-Pass} & {Parameters} & {mIoU} & {mAcc}\\
\midrule
3 &  5 & 12.10 ms & $402 \cdot 10^3$ & 50.04 & 57.42\\
3 & 10 & 13.64 ms & $402 \cdot 10^3$ & 50.98 & 58.16\\
3 & 20 & 17.65 ms & $402 \cdot 10^3$ & \textbf{52.25} & \textbf{60.83}\\
\midrule
5 &  5 & 14.53 ms & $625 \cdot 10^3$ & 52.69 & 58.87\\
5 & 10 & 17.12 ms & $625 \cdot 10^3$ & 52.91 & 59.57\\
5 & 20 & 23.35 ms & $625 \cdot 10^3$ & \textbf{53.27} & \textbf{60.15}\\
\midrule
7 &  5 & 16.99 ms & $880 \cdot 10^3$ & 52.93 & 59.87\\
7 & 10 & 20.68 ms & $880 \cdot 10^3$ & 53.57 & 60.92\\
7 & 20 & 29.38 ms & $880 \cdot 10^3$ & \textbf{53.93} & \textbf{61.73}\\
\bottomrule
\end{tabular}
}
\label{tab:segmentation_results_stanford_depth}   
\end{table}

By increasing the number of convolutional layers, we can build deeper networks.
Similar to discrete convolutions, deep point convolutional networks perform better than shallow ones.
Equally, the performance increases with the number of neighbors.
However, increasing the number of neighbors increases the computational cost, resulting in slower inference times.
Furthermore, increasing the number of convolutions leads to additional memory consumption.

\parag{Dilation Factor \textit{d} (\reftab{segmentation_results_stanford_dilation}).}
\begin{table}[t]
\center
\vspace{5px}
\caption{Ablation Study: Dilated Point Convolutions. Varying dilation factors $d$. Dataset: S3DIS Area 5.
}
\resizebox{0.99\columnwidth}{!}
{%
\setlength{\tabcolsep}{3pt}
\begin{tabular}{ccccccc}
\toprule 
{Number of}   & {Number of}     & {Time per}     & {Number of}  & {Dilation} & &\\
{PointConvs} & {Neighbors $k$} & {Forward-Pass} & {Parameters}& $d$ & {mIoU} & {mAcc}\\
\midrule
7 &  20 & 29.38 ms & $880 \cdot 10^3$ &1 & 53.93 & 61.73\\
7 &  20 & 31.57 ms & $880 \cdot 10^3$ &2 & 55.83 & 61.76\\
7 &  20 & 35.36 ms & $880 \cdot 10^3$ &8 & \textbf{61.28} & \textbf{68.38}\\
7 &  20 & 51.65 ms & $880 \cdot 10^3$ &16 & 58.79 & 65.84\\
\bottomrule 
\vspace{-17px}
\end{tabular}
}
\label{tab:segmentation_results_stanford_dilation}    
\end{table}
Dilated Point Convolutions are an efficient tool to rapidly increase the receptive field of convolutions.
Using dilation, the receptive field can be increased significantly (\reffig{receptive_field}) at constant memory requirements and a marginal increment in processing time.
The improved performance on the semantic segmentation task shows that indeed a larger receptive field is important.
However, the rapidly increasing receptive field resulting in large receptive fields in later layers is also responsible for sparsely sampled neighborhoods in earlier layers.
We assume that this makes it harder for the network to learn high-frequency or local features.
In future work, it could be interesting to investigate deep convolutional networks using Dilated Point Convolutions with a dilation rate $d$ that increases with the depth of the network.
Intuitively, such a network could learn localized signals in the earlier stages and higher level information at later stages.

\parag{Model Size.}
Note that, since the kernel function $g(p)$ is defined over relative point positions $p$, the number of trainable parameters is independent of the number of neighbors $k$ (and hence the dilation factor $d$). As such, increasing the number of neighbors $k$ (or the dilation factor $d$) increases the receptive field without increasing the model size.

\section{Conclusion}
\label{sec:conclusion}
In this work, we reviewed several mechanisms to increase the receptive field size of 3D point convolutions. We analyzed and compared different network architectures based on the receptive field size which we showed to be directly related to the performance of point convolutional networks. 
Specifically, we have proposed \emph{dilated point convolutions} as an elegant and efficient technique to significantly increase the receptive field size of point convolutions. 
As a result, we were able to report solid improvements over well-known baseline methods for 3D semantic segmentation and 3D object classification.
More importantly, our dilation mechanism can easily be integrated into most existing point convolutional networks.
We hope these insights enable the research community to develop better performing models.

\textbf{Acknowledgements:} 
This work was supported by the ERC Consolidator Grant DeeViSe(ERC-2017-COG-773161).
We thank Mats Steinweg, Dan Jia, Jonas Schult and Alexander Hermans for their valuable feedback.

\newpage

\bibliographystyle{plain}
\balance
\bibliography{abbreviations,egbib}

\end{document}